\documentclass[conference, final]{IEEEtran}
\usepackage[utf8]{inputenc}
\usepackage{multirow}
\usepackage{caption}
\usepackage{xcolor, soul}
\usepackage{amsmath, amssymb}
\usepackage{tcolorbox}
\usepackage{subcaption}
\usepackage{graphicx}
\usepackage{tabularx}
\usepackage{booktabs}
\usepackage{placeins}[subsection]

\usepackage[left=1.62cm, right=1.62cm, top=0.75in, bottom=1.06in]{geometry}
\usepackage{verbatim}

\title{Investigating Distributions of Telecom Adapted Sentence Embeddings for Document Retrieval}

\author{\IEEEauthorblockN{Sujoy Roychowdhury, Sumit Soman, Ranjani Hosakere Gireesha, Vansh Chhabra\IEEEauthorrefmark{1}, Neeraj Gunda\IEEEauthorrefmark{1}, \\
Subhadip Bandyopadhyay, Sai Krishna Bala}
\IEEEauthorblockA{Ericsson R\&D, Bangalore\\
Email: \{sujoy.roychowdhury, sumit.soman, ranjani.h.g, subhadip.bandyopadhyay, sai.krishna.bala\}@ericsson.com
}
    \IEEEcompsocitemizethanks{
        \IEEEcompsocthanksitem \IEEEauthorrefmark{1} This work was done by the authors during their internship at Ericsson.
    }
        \thanks{
        \IEEEcompsocthanksitem Accepted for the \textit{Workshop On Next Gen Networks Through LLMs Action Models and Multi Agent Systems} at ICC 2025.
    }
    }

\PassOptionsToPackage{hyphens}{url}\usepackage{hyperref}
\IEEEoverridecommandlockouts

\begin{document}

\maketitle

\begin{abstract}
A plethora of sentence embedding models makes it challenging to choose one, especially for technical domains rich with specialized vocabulary. In this work, we domain adapt embeddings using telecom data for question answering. We evaluate embeddings obtained from publicly available models and their domain-adapted variants, on both point retrieval accuracies, as well as their (95\%) confidence intervals.  We establish a systematic method to obtain thresholds for similarity scores for different embeddings. As expected, we observe that fine-tuning improves mean bootstrapped accuracies. We also observe that it results in tighter confidence intervals, which further improve when pre-training is preceded by fine-tuning. We introduce metrics which measure the distributional overlaps of top-$K$, correct and random document similarities with the question. Further, we show that these metrics are correlated with retrieval accuracy and similarity thresholds. Recent literature shows conflicting effects of isotropy on retrieval accuracies. Our experiments establish that the isotropy of embeddings (as measured by two independent state-of-the-art isotropy metric definitions) is poorly correlated with retrieval performance. We show that embeddings for domain-specific sentences have little overlap with those for domain-agnostic ones, and fine-tuning moves them further apart. Based on our results, we provide recommendations for use of our methodology and metrics by researchers and practitioners.
\end{abstract}

\section{Introduction} \label{sec:intro}

Question Answering (QA) methods such as Retrieval Augmented Generation (RAG) typically involve retrieval of sections, paragraphs or sentences from a document corpus to accurately answer user queries. Embedding models are used to map the questions or documents to a semantic space. Retrieval is typically achieved by computing similarity between embeddings of questions and those of documents. The most similar top-$K$ documents are considered to be relevant. 

Although many state-of-the-art (SOTA) models trained on publicly available datasets are accessible \cite{reimers2019sentence, chen2024bge, llm_embedder,bge_embedding}, obtaining good retrieval accuracies for domain-specific tasks is challenging \cite{roychowdhury2024evaluation}. It is well acknowledged in the literature that domain adaptation and fine-tuning can  improve retrieval \cite{li2020sentence}, but making an informed choice among several available models involves extensive evaluation over parameters such as the number of relevant documents retrieved for a test set.


Some studies \cite{zhou2022problems} have identified limitations of cosine similarities in retrieving embeddings: a sample limitation is an underestimation of the similarity of frequent words with their homonyms. It has been shown that cosine similarities can be arbitrary or dependent on regularization, making them unreliable for retrieval tasks \cite{steck2024cosine} - although this study was limited to linear models the authors have conjectured that the same may be true for non-linear models. In fact, variations in embedding space representations obtained from different architectures have been widely studied \cite{mistry2023comparative, bis2021too, timkey2021all}. Another limitation observed is reporting of point accuracies, without any error bars, for retrieval tasks. This limits estimation of performance on new questions, especially when evaluated with relatively small datasets.  

Recent work has explored isotropy as a measure for quantifying robust embedding space representations \cite{jung2023isotropic, rudman2023stable, rudman2021isoscore}, though it has also been argued otherwise \cite{hou2024anisotropic, ait2023anisotropy, godey2023anisotropy, razzhigaev2023shape}. In particular, \cite{jung2023isotropic} suggests that isotropic embeddings improve retrieval whereas \cite{rudman2023stable} propose that reduced isotropy or anisotropy helps retrieval. \cite{rajaee2021does} looks at isotropy of embeddings and show that increasing the isotropy of fine-tuned models leads to poorer performance.

\color{black}
We observe a few limitations with the current practice of measuring retrieval performance in both research and practice. First, reporting point accuracies do not provide insight into error bars (confidence intervals). This is especially important for relatively smaller datasets. Second, the lack of confidence intervals does not allow for tests of statistical significance when comparing different embedding models or domain adaptation strategies. Third, to the best of our knowledge, we have not found prior work which has provided a systematic approach to choose the best threshold. In practice, such thresholds are often chosen by inspection of similarity scores. Our approach of bootstrapping provides the ability to perform tests for statistical significance on the results, and we choose the maximum threshold such that our results are not statistically worse off. Finally, although prior work \cite{gao2021simcse, ethayarajh2019contextual} have looked at the effect of domain adaptation on embeddings, the separation of domain-specific embeddings from general purpose embeddings under domain adaptation has not been studied. This does not allow a clear understanding of why performance changes on general purpose retrieval post domain adaptation.
\color{black}

\subsection{Research Questions and Contributions}

The primary research questions in this work are as follows:

\begin{itemize}
    \item \textbf{RQ1}: What are the confidence intervals (CI) of accuracies of SOTA retrieval models and their fine-tuned versions when considering telecom-specific tasks?
    \item \textbf{RQ2}: What facets apart from retrieval accuracies can characterize an embedding model? How does the distribution of cosine similarities vary across emwbeddings?
    \item \textbf{RQ3}: Can the variation of retrieval accuracies be attributed to only the isotropy of the embeddings? 
\end{itemize}


Our \textbf{primary contributions} are:
\begin{itemize}
    \item Demonstrate that fine-tuning improves accuracy and CI. Pre-training before fine-tuning improves CI further. 
    \item Propose a systematic method to introduce thresholds with minimal effect on retrieval accuracies.
    \item {\color{black}{Show}} that although domain adaptation via fine tuning leads to higher isotropy scores, retrieval performance across models is poorly correlated with the isotropy scores of the models; improving isotropy scores via transformations does not improve accuracies.
    \item We introduce metrics which measure the distributional overlaps of top-$K$, correct and random document similarities with the question. 
    \item Show empirically that these metrics are correlated with accuracies and similarity thresholds.
    \item Demonstrate that domain adaptation shifts the embeddings of the target domain further away from embeddings of sentences from domain-agnostic datasets.
\end{itemize}

The rest of the paper is structured as follows: the methodology is detailed in Section \ref{sec:methodology}. We describe the telecom dataset and embedding models in Section \ref{sec:Datasets} and Section \ref{subsec:embeddingModels} respectively. We report experimental results of multiple embeddings (with and without domain adaptation) in Section \ref{subsec:results}. We summarize our findings and discuss the limitations and scope of future work in Section \ref{sec:conclusions}.

\section{Methodology}\label{sec:methodology}

In this study we consider the following: computing bootstrapped accuracies, estimating probabilities of overlap between different distributions, analysis of minimum thresholds for similarities and study the effects of isotropy scores. We describe each of these formally in this section. For most of our experiments, we choose a bootstrapped approach to get both point estimates and CI for our estimates. 

Consider a dataset $\mathcal{D} = [s_1, s_2, \ldots, s_N]$, where $s_i$ is the $i^{th}$ sentence and $i \in [1,N]$. Let $\mathcal{D}$  be associated with a question set $\mathcal{Q}$, containing $Q$ questions. Each question $q \in \mathcal{Q}$ can be uniquely answerable by one sentence $s_q \in \mathcal{D}$, which we consider as the correct answer for the question $q$. Let the embedding representation of $s_i$ using a sentence embedding model $\mathcal{M}$ be represented by $E_\mathcal{M}(s_i)$, and correspond to  dimension $\mathcal{M}_p$. Similarly, let $E_\mathcal{M}(q)$ represent the embedding (using sentence embedding model $\mathcal{M}$) for a question, $q \in \mathcal{Q}$. Henceforth, in this work, all sentence embeddings will be referred to as embeddings. 

Like in any typical QA retrieval methodology, $\mathcal{D}$ and $\mathcal{Q}$ result in embedding matrices of sizes $N \times \mathcal{M}_p$ and $Q \times \mathcal{M}_p$ respectively. All embeddings are normalized to have unit $L_2$ norm. We draw $m$ bootstrap samples from $\mathcal{Q}$, each containing $l$ questions i.e., $|\mathcal{Q}_j| = l$ with $|\cdot|$ indicative of the cardinality of the corresponding set and $j \in [1,m]$. We use these bootstrapped samples in our experiments.

\subsection{Bootstrapped metrics}\label{subsec:bootstrapAcc}

 Consider any $j^{th}$ bootstrap sample $\mathcal{Q}_j \in \mathcal{Q}$. For each question $q \in \mathcal{Q}_j$, we find the set $t_q^K$ of the top-$K$ most similar sentences based on highest cosine similarity and check if $s_q$ is included in this set. The top-$K$ accuracy, $a_j$, is the proportion of questions in this bootstrap sample for which $s_q \in t_q^K$. The mean bootstrapped retrieval accuracy is given by $a = \frac{1}{m} \sum_{j=1}^{m} a_j$. 

The $95\%$ confidence interval $(a_{\text{lower}}, a_{\text{upper}})$ is defined by the $2.5^{th}$ and $97.5^{th}$ percentiles of the set of $a_i$ values. {\color{black} This approach is not limited to computing accuracies alone, but can be replicated for other relevant metrics like Normalized Discounted Cumulative Gain (NDCG).}

\subsection{Computation of thresholds}\label{subsec:thresholds}

It is often desirable to have thresholds on similarity scores between questions embeddings and retrieved sentence embeddings from the dataset via top-$K$ similarity scores, thus ignoring any sentence with similarity score below this threshold. This reduces retrieval of sentences that may not necessarily answer the question. A low threshold runs the risk of including wrong/irrelevant documents in retrieval results, and a high threshold can reduce the top-$K$ accuracy.

However, there is no reliable way to estimate a threshold, given that the distribution of similarities can be different based on choice of the embedding model. Hence, we follow a bootstrapped analysis. Consider each of the bootstrap samples, $\mathcal{Q}_j$. We construct a similarity matrix $S_\mathcal{M}^j=E_\mathcal{M}(\mathcal{Q}_j)\cdot E_\mathcal{M}(\mathcal{D})^T$, where $(\cdot)$ denotes the dot product, $()^T$ denotes the matrix transpose and $S_\mathcal{M}^j \in \mathbb{R}^{(l \times N)}$. 
Let $T_\mathcal{M}^j$ be constructed such that, each row of $T_\mathcal{M}^j$ has the top-$K$ similarity scores from $S_\mathcal{M}^j$. 
We define $\gamma^{j} = min(T_\mathcal{M}^{j})$ and $\Gamma \triangleq \{ \gamma^{j}: j \in [1,m]\}$. {\color{black} This choice of $\gamma_j$ ensures that if the threshold is set to be lower than $\gamma_j$ then the performance on bootstrap $j$ is unaffected since all similarity scores will remain untouched in $T_\mathcal{M}^j$.}

\begin{figure}[hbtp]
\centering
\includegraphics[width=0.9\columnwidth]{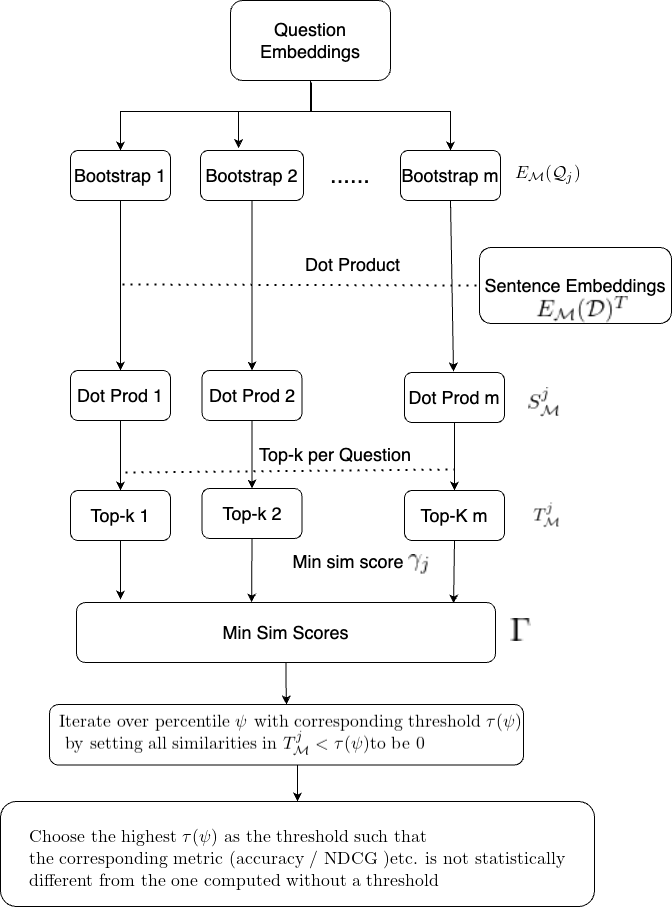}
    \caption{Schematic diagram of threshold determination using $m$ bootstraps, the index $j$ going from $1$ to $m$}
    \label{fig:thresholdDetermination}
\end{figure}

We choose a threshold, using $\psi^{th}$ percentile of $\Gamma$, defined by $\tau(\psi)$ s.t. $P_\Gamma (x < \tau(\psi)) = \psi$. We study the effect of $\tau(\psi)$ on bootstrapped retrieval accuracies. We substitute all similarities of $T_\mathcal{M}^j < \tau(\psi)$ to be zero. We consider the threshold as the highest $\tau(\psi)$ such that the {\color{black} metric e.g. accuracy / NDCG}  from this substitution is not statistically different from the mean bootstrap accuracy, $a$ (refer Section \ref{subsec:bootstrapAcc}).  We clarify that $\gamma^{j}$ is the set of minimum similarities in the bootstrapped samples, thus $\psi$ can be interpreted as the percentile of irrelevant documents - however, there is no direct interpretation with respect to the total number of documents retrieved. {\color{black} The process for threshold determination is also shown as a schematic diagram in Fig. \ref{fig:thresholdDetermination}}

{\color{black} We note that our approach ensures that the obtained metric (accuracy / NDCG etc.) is not statistically different from one without a threshold - this feature is possible to be ensured only because we have followed bootstrapping and thus getting the capability to do statistical testing. We also observe that thresholding can either keep accuracy same or reduce it. On the other hand a metric like NDCG will offer a tradeoff with ranked position as well as fewer documents retrieved. In both cases however our approach ensures performance does not degrade in a statistical sense.}



\subsection{Analysis of distribution of vector embeddings}\label{subsec:overlap}

To understand vector embeddings in the semantic space and their effect on retrieval, we study distributions of cosine similarities of embeddings from selected models. As mentioned earlier, all embeddings have unit $L_2$ norm. We first consider $\mathcal{Q}$ and estimate the following distributions:
\begin{itemize}
    \item \textbf{Distribution of correct similarity scores} - Let $sim^{corr}_q$ represent the cosine similarity between $E_\mathcal{M}(q)$ and $E_\mathcal{M}(s_q)$, $\forall q \in \mathcal{Q}$. Let $S_{corr}=\{sim^{corr}_q:q \in \mathcal{Q}\}$ represent the set of correct similarity scores. 
    \item \textbf{Distribution of top-k similarity scores} - Let $sim^{topK}_q$ represent cosine similarities between any question and the corresponding top-$K$ retrieved sentences.  
    Let this set be represented by $S_{topK}=\{sim^{topK}_q:q \in \mathcal{Q}\}$. 
    \item \textbf{Distribution of random similarity scores} - Let $sim^{rand}_q$ represent the cosine similarity between embedding of any question, $E_\mathcal{M}(q)$, $\forall q \in \mathcal{Q}$ and that of a randomly chosen statement $E_\mathcal{M}(s_r)$, s.t. $s_r \in \mathcal{D}$. Let this set be represented by $S_{rand}=\{sim^{rand}_q:q \in \mathcal{Q}\}$.  
\end{itemize}
Evidently, $|S_{corr}|=Q$, $|S_{topK}|=KQ$ and $|S_{rand}|=Q$.

We estimate the Empirical Cumulative Distribution Function (ECDF) 
for each of these sets; let these be  $C_{corr}$, $C_{topK}$ and $C_{rand}$ for $S_{corr}$, $S_{topK}$ and $S_{rand}$ respectively. 




Consider each bootstrapped sample $\mathcal{Q}_j$. Let $\theta_j$ be the similarity score at the $\psi^{th}$ percentile of the set $S_{topK}$ i.e., $P_{S_{topK}}(sim^{topK} \le \theta_j)=\psi$. Now, we define the following ECDF estimates: 
\begin{align}
C_{corr}({\theta_j}) \triangleq P_{S_{corr}}(sim^{corr} > \theta_j) 
\\
C_{rand}({\theta_j}) \triangleq P_{S_{rand}}(sim^{rand} > \theta_j) \label{eqn:correctTopki}
\end{align}

These are a measure of the overlap of cosine similarities between top-$K$ and correct, top-$K$ and random QA sentence pairs. 
The mean of these across the bootstrapped samples can be calculated as $\bar{C}_{corr}(\theta)=\frac{1}{m}\sum_{j=1}^{m} C_{corr}({\theta_j})$ and $\bar{C}_{rand}(\theta)=\frac{1}{m}\sum_{j=1}^{m} C_{rand}({\theta_j})$. We refer to them as correct-overlap-ECDF (COE) and random-overlap-ECDF (ROE) estimates. We also estimate the $95\%$ CI for both COE and ROE by the using the $2.5^{th}$ and $97.5^{th}$ percentile of  $C_{corr}({\theta_j})$ and $C_{rand}({\theta_j})$ as lower and upper bounds respectively.


\subsection{Domain Adaptation}\label{subsec:DomainAdapt}

One of the key challenges in leveraging embedding models for technical domains is the lack of domain specific knowledge, since the SOTA (base) models have been trained on publicly available datasets which may be minimally introduced to domain specific terminology. We evaluate various domain adaptation techniques on the base models:
\begin{itemize}
    \item Pre-training \cite{li2020sentence}: We use Masked Language Modeling (MLM) \cite{salazar2019masked} approach for this. Sentences from the corpus of technical documents (of a domain) are used.
    \item  Fine-tuning \cite{mosbach2020interplay}: We prepare triplets of the form $<q,p,n>$ where $q$ corresponds to the user query, $p$ represents the correct (positive) answer and $n$ is a list of incorrect (negative) answers. The base model is fine-tuned using these triplets. It may be noted here that the fine-tuning may be performed post pre-training or independently on the base model (without pre-training). 
\end{itemize}
Thus, we evaluate the following variants of embedding models - base model, pre-trained only (PT), fine-tuned only (FT) and pre-training followed by fine-tuning (PT-FT). Post fine-tuning, we merge the base model with the domain adapted model.

\subsection{Isotropy Scores}\label{subsec:isotropyScores}


Isotropy measures distribution of embeddings on the high-dimensional unit hypersphere (since all embeddings have unit-$L_2$ norm). If the embeddings are uniformly distributed over the unit sphere i.e. there is no preferred direction, then, they are said to be isotropic \cite{arora2016latent, mu2017all}. We use two different measures of isotropy to validate our findings. We represent the isotropic scores as, $I_A$, the second order approximation as defined in \cite{mu2017all} and $I_B$ to be isoscores as per \cite{rudman2021isoscore,rudman2022isoscore}. These measure isotropy differently and thus their scores can be quite different. Higher isotropic scores implies embeddings being well distributed in the unit hyper-sphere.

Various transformations have been proposed in literature to improve isotropy scores. We choose the following to study the effect of isotropy (measured using both $I_A$, $I_B$):
\begin{itemize}
    \item Whitened: Whitening of embeddings \cite{jung2023isotropic}
    \item PCA: Post-processing embeddings by centering and eliminating the top principal components \cite{mu2017all} 
    \item Standardized: Mean subtraction and unit std. dev. \cite{timkey2021all} 
\end{itemize}



\subsection{Comparison of Embeddings Post Domain Adaptation} \label{subsec:embeddingDomainAdapt}

We analyze the effect of pre-training and fine-tuning base embedding models with domain-specific data by comparing distribution of the resultant embeddings with that of embeddings from a domain-agnostic dataset. 

Let $\mathcal{D}$ represent domain-specific data, $\mathcal{D}'$ represent domain-agnostic dataset. Let $\mathcal{M}$ be the base model, $\mathcal{M}'$ be the pre-trained, fine-tuned version of the base model. Let similarity between the datasets be defined 
$\Delta_\mathcal{M}(\mathcal{D},\mathcal{D}') \triangleq \{ min (|| E_\mathcal{M}(d), E_\mathcal{M}(d')||_2) : d \in \mathcal{D}, d' \in \mathcal{D}'\}$, and $|\Delta_\mathcal{M}(\mathcal{D}, \mathcal{D}')| = |\mathcal{D}|$. 

We compare the distributions of $\Delta_{\mathcal{M}}$ and $\Delta_{\mathcal{M}'}$. Our motivation here is to analyse the separation of the distributions post domain adaptation. 

\section{Experimental setup}
\subsection{Datasets} \label{sec:Datasets}

Our primary domain specific dataset, $\mathcal{D}$, is an internal dataset for domain-specific QA. This has been curated by Subject Matter Experts (SME) and consists of sections from 3GPP specifications Release 17 \cite{3gpp_release_17}. The dataset consists of  5167 questions from 452 paragraphs/contexts. These paragraphs constitute total of 5257 sentences; NLTK's sentence tokenizer is used for extracting sentences \cite{loper2002nltk}. Training and test split considered is 80\% and 20\% respectively. 

\subsection{Embedding Models}\label{subsec:embeddingModels}

We consider the following embedding models:
\begin{itemize}
    \item From BAAI, we consider \textit{bge-large-en} \cite{bge_embedding} and \textit{llm-embedder} \cite{llm_embedder} with $\mathcal{M}_p = 1024$, $768$ respectively. We PT, FT, PT-FT these models for further experiments. 
    \item In addition, only for the telecom dataset
    \begin{itemize}
        \item We evaluate a telecom adapted BERT model General-Telecom-Embeddings (GTE), $\mathcal{M}_p = 768$.
        \item From the sentence transformers 
    \cite{reimers2019sentence} library, we consider MPNET \cite{song2020mpnet} and MiniLM (\textit{all-MiniLM-L6-v2}). {Their $\mathcal{M}_p$ are 768 and 384 respectively}.
     \item From OpenAI family\footnote{\url{https://platform.openai.com/docs/guides/embeddings/embedding-models}}, we evaluate \textit{text-embedding-3-small}, \textit{text-embedding-3-large} and \textit{ada\_002}, for $\mathcal{M}_p=$ 1536, 3072 and 1536 respectively.
    \end{itemize}
\end{itemize} 

All experiments used a A100-SXM4-80GB GPU. 
\section{Results}
\label{subsec:results}

\subsection{Accuracies and Confidence Intervals}\label{subsec:accs}

Table \ref{tab:acc_ndcg}  reports retrieval accuracy along with confidence interval widths.  We observe consistent accuracy improvements across models on FT and PT-FT. However, we observe that fine-tuning a base model and that of a pre-trained model is not much different from the mean accuracies. More importantly, and to the best of our knowledge not reported previously, is the insight that confidence intervals become tighter with FT and further, with PT-FT. Since only PT is trained with a MLM objective, it is not surprising, and previously observed \cite{li2020sentence}, that there is a reduction in accuracies for PT models. {\color{black} We also shows the bootstrapped NDCG scores and the width of the confidence interval. We observe that even for NDCG, the width of the confidence interval also reduces with domain adaptation, especially for PT-FT models.  Table \ref{tab:acc_ndcg} also has the accuracies and NDCG for the full dataset without bootstrapping.}

We report COE (as defined in Section \ref{subsec:overlap}) for the various models and domain-specific datasets in  Table \ref{tab:acc_ndcg}. The correlation between COE and accuracy is reported in Table. \ref{tab:corr}. We see a strong positive correlation between them.

The column $\tau({\psi})$ in Table \ref{tab:acc_ndcg} indicates the thresholds as per the method described in Section \ref{subsec:thresholds}. While the accuracies have slightly reduced with introduction of thresholds (refer Acc @$\tau$ column), this can be interpreted as the accuracy obtained with 
removal of less relevant documents in retrieved results.  Additionally, Acc @$\tau$ \textbf{is not statistically different} from the bootstrapped accuracy for the whole dataset (refer column 7 vs column 2). Thus, our choice of threshold \textbf{does not lead to degradation of accuracies} in a statistical sense. We re-iterate that there is no direct interpretation of $\psi$ with respect to the total number of documents retrieved.

\begin{figure*}[hbtp]
\centering
\includegraphics[scale=0.25]{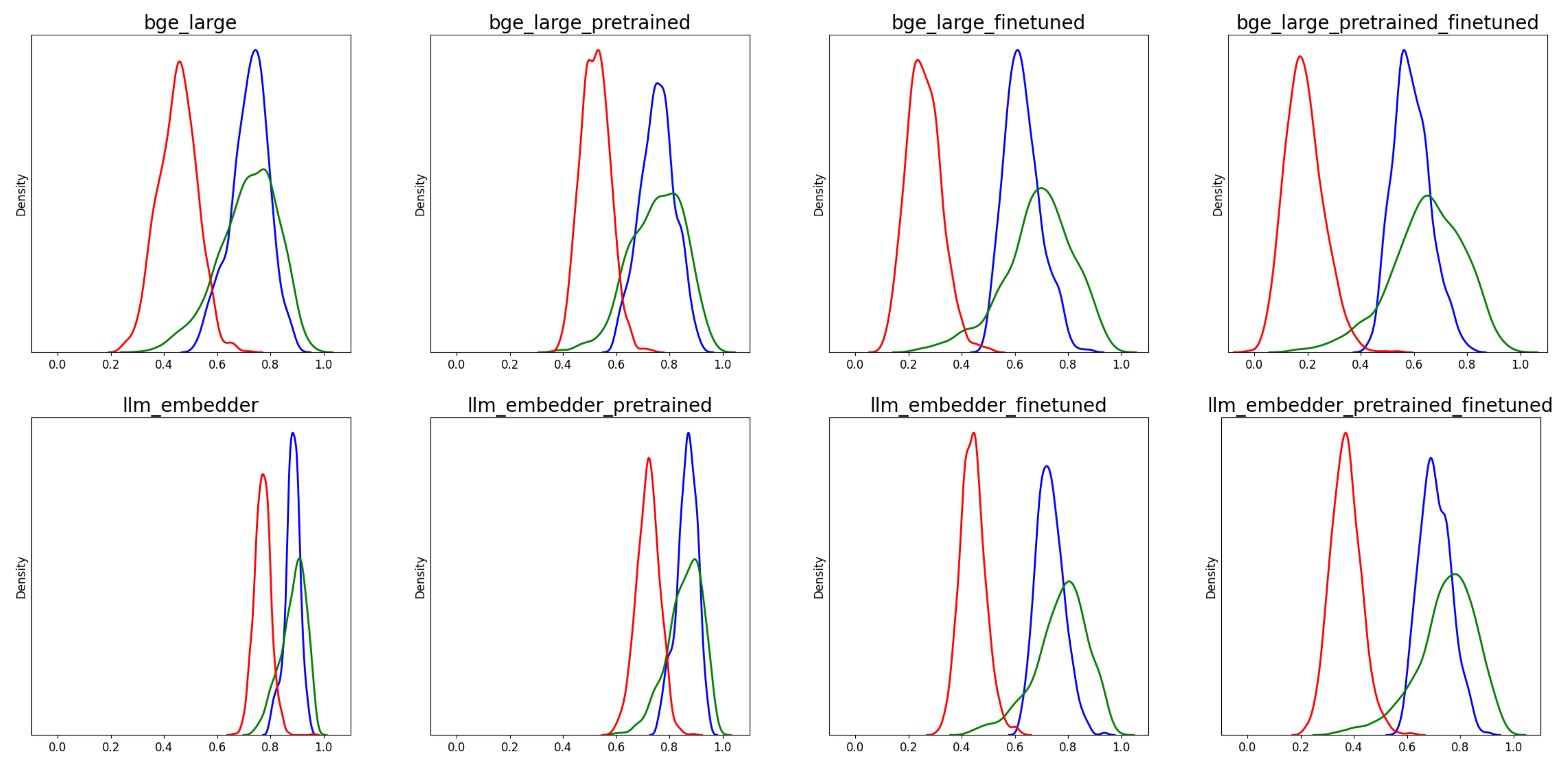}
    \caption{Density plots for telecom dataset. Red, green and blue indicate distribution of $S_{rand}$, $S_{corr}$ and $S_{topK}$ respectively. Refer Sec. \ref{subsec:overlap} for definitions}
    \label{fig:hists_TelecomQuad}
\end{figure*}

As expected, the correlation between ROE and accuracy is low (refer Table \ref{tab:corr}) across domains. 
We analyze the correlation between threshold ($\tau(\psi)$) with ROE. This is found to be positively correlated. These correlations are not obvious - this indicates that for a model to perform well, questions must be well interspersed with answers in the embedding space. This is also reflected in the distribution of embeddings in Figure \ref{fig:hists_TelecomQuad}. 

On further analysing Figure  \ref{fig:hists_TelecomQuad}, we notice that the \textit{llm\_embedder} model has a very peaky distribution of cosine similarities (even for $S_{rand})$. This is indicative of a model with low isotropy. Despite being less isotropic, the retrieval accuracies of the model is similar to the \textit{bge\_large} model which is more isotropic. The domain adaptation of \textit{llm\_embedder} model creates a wider distribution of the cosine similarities indicating better isotropy. The improvement in isotropy post domain-adaptation has also been reported in \cite{gao2021simcse}.

\subsection{Isotropy Score Analysis}\label{subsec:isotropyResults}
Table \ref{tab:transformations} lists the retrieval accuracies for the telecom dataset $\mathcal{D}$, isotropic measures $I_A$ and $I_B$ of base and adapted models for various transformations (intended to increase isotropy scores and described in Section \ref{subsec:isotropyScores}).


\begin{table*}[hbtp]
\centering
\scalebox{0.8}{

\begin{tabular}{|c|l|l|l|l|l|}
\hline
\textbf{Corr}    & Acc v. COE                & Acc v. ROE                 & Thresh v. ROE           & Acc v. $I_A$               & Acc v. $I_B$              \\ \hline
\textbf{Telecom} & \multicolumn{1}{r|}{0.882} & \multicolumn{1}{r|}{-0.121} & \multicolumn{1}{r|}{0.391} & \multicolumn{1}{r|}{0.014} & \multicolumn{1}{r|}{0.05} \\ \hline
\end{tabular}
}
\caption{Correlation values}\label{tab:corr}
\end{table*}
\begin{table*}[hbtp]
\centering
\scalebox{0.82}{
\begin{tabular}{|l|llllllll|ll|}
\hline
\multicolumn{1}{|c|}{}                & \multicolumn{8}{c|}{Bootstrapping}                                                                                                                                                                                                                & \multicolumn{2}{c|}{Baseline (full data)}             \\ \hline
\multicolumn{1}{|c|}{Embedding Model} & \multicolumn{1}{c|}{Acc}   & \multicolumn{1}{l|}{Acc-CI}    & \multicolumn{1}{c|}{NDCG} & \multicolumn{1}{l|}{NDCG-CI}  & \multicolumn{1}{c|}{COE}   & \multicolumn{1}{c|}{ROE}  & \multicolumn{1}{c|}{$(\tau, \psi)$} & \multicolumn{1}{c|}{Acc @ $\tau$} & \multicolumn{1}{c|}{Acc}  & \multicolumn{1}{c|}{NDCG} \\ \hline
bge\_large                            & \multicolumn{1}{l|}{66.87} & \multicolumn{1}{l|}{17.04} & \multicolumn{1}{l|}{29.6} & \multicolumn{1}{l|}{0.6} & \multicolumn{1}{l|}{87.98} & \multicolumn{1}{l|}{4.81} & \multicolumn{1}{l|}{0.5 (35)}       & 67.18                             & \multicolumn{1}{l|}{66.0} & 29.9                      \\ 
bge\_large\_pretrained                & \multicolumn{1}{l|}{62.64} & \multicolumn{1}{l|}{17.0}  & \multicolumn{1}{l|}{27.2} & \multicolumn{1}{l|}{0.4} & \multicolumn{1}{l|}{85.94} & \multicolumn{1}{l|}{2.18} & \multicolumn{1}{l|}{0.58 (25)}      & 61.36                             & \multicolumn{1}{l|}{63.1} & 27.5                      \\ 
bge\_large\_finetuned                 & \multicolumn{1}{l|}{81.61} & \multicolumn{1}{l|}{14.04} & \multicolumn{1}{l|}{34.2} & \multicolumn{1}{l|}{1.2} & \multicolumn{1}{l|}{91.98} & \multicolumn{1}{l|}{0.22} & \multicolumn{1}{l|}{0.43 (25)}      & 79.46                             & \multicolumn{1}{l|}{82.0} & 34.2                      \\ 
bge\_large\_pretrained\_finetuned     & \multicolumn{1}{l|}{81.67} & \multicolumn{1}{l|}{13.04} & \multicolumn{1}{l|}{34.9} & \multicolumn{1}{l|}{0.5} & \multicolumn{1}{l|}{91.06} & \multicolumn{1}{l|}{0.23} & \multicolumn{1}{l|}{0.4 (35)}       & 77.73                             & \multicolumn{1}{l|}{81.5} & 34.9                      \\ \hline
llm\_embedder                         & \multicolumn{1}{l|}{70.06} & \multicolumn{1}{l|}{14.52} & \multicolumn{1}{l|}{29.2} & \multicolumn{1}{l|}{1.6} & \multicolumn{1}{l|}{87.26} & \multicolumn{1}{l|}{5.77} & \multicolumn{1}{l|}{0.78 (30)}      & 69.9                              & \multicolumn{1}{l|}{69.2} & 29.3                      \\ 
llm\_embedder\_pretrained             & \multicolumn{1}{l|}{57.12} & \multicolumn{1}{l|}{19.57} & \multicolumn{1}{l|}{25.2} & \multicolumn{1}{l|}{0.8} & \multicolumn{1}{l|}{84.88} & \multicolumn{1}{l|}{6.32} & \multicolumn{1}{l|}{0.75 (30)}      & 52.53                             & \multicolumn{1}{l|}{57.0} & 25.2                      \\ 
llm\_embedder\_finetuned              & \multicolumn{1}{l|}{81.58} & \multicolumn{1}{l|}{13.52} & \multicolumn{1}{l|}{34.3} & \multicolumn{1}{l|}{0.6} & \multicolumn{1}{l|}{90.73} & \multicolumn{1}{l|}{0.10} & \multicolumn{1}{l|}{0.56 (40)}      & 80.69                             & \multicolumn{1}{l|}{81.8} & 34.4                      \\ 
llm\_embedder\_pretrained\_finetuned  & \multicolumn{1}{l|}{80.37} & \multicolumn{1}{l|}{12.52} & \multicolumn{1}{l|}{33.7} & \multicolumn{1}{l|}{0.5} & \multicolumn{1}{l|}{90.74} & \multicolumn{1}{l|}{0.21} & \multicolumn{1}{l|}{0.53 (25)}      & 77.97                             & \multicolumn{1}{l|}{80.8} & 33.8                      \\ \hline
\end{tabular}
}
\caption{Performance metrics using bootstrapping compared to baseline on full dataset. CI - width of confidence interval.}
\label{tab:acc_ndcg}
\end{table*}
\begin{table*}[h]
\centering
\scalebox{0.82}{
\begin{tabular}{|lllllllll|}
\hline
\multicolumn{1}{|c|}{\textbf{Embedding Model}}                      & \multicolumn{2}{l|}{\textbf{Baseline}}                                  & \multicolumn{2}{c|}{\textbf{Standardized}}                              & \multicolumn{2}{c|}{\textbf{Whitened}}                                  & \multicolumn{2}{c|}{\textbf{PCA}}                  \\ \hline
\multicolumn{1}{|l|}{}                                     & \multicolumn{1}{l|}{Acc}   & \multicolumn{1}{l|}{$I_A, I_B$}   & \multicolumn{1}{l|}{Acc}   & \multicolumn{1}{l|}{$I_A, I_B$}   & \multicolumn{1}{l|}{Acc}   & \multicolumn{1}{l|}{$I_A, I_B$}   & \multicolumn{1}{l|}{Acc}   & $I_A, I_B$   \\ \hline
\multicolumn{1}{|l|}{bge\_large}                           & \multicolumn{1}{l|}{66.87} & \multicolumn{1}{l|}{9.24, 27.81}  & \multicolumn{1}{l|}{66.63} & \multicolumn{1}{l|}{9.71, 97.23}  & \multicolumn{1}{l|}{65.11} & \multicolumn{1}{l|}{9.41, 79.15}  & \multicolumn{1}{l|}{68.43} & 16.91, 95    \\ 
\multicolumn{1}{|l|}{bge\_large\_pretrained}               & \multicolumn{1}{l|}{62.64} & \multicolumn{1}{l|}{6.34, 23.77}  & \multicolumn{1}{l|}{59.24} & \multicolumn{1}{l|}{6.82, 96.26}  & \multicolumn{1}{l|}{63.17} & \multicolumn{1}{l|}{6.78, 24.96}  & \multicolumn{1}{l|}{57.02} & 12.36, 92.75 \\ 
\multicolumn{1}{|l|}{bge\_large\_finetuned}                & \multicolumn{1}{l|}{81.61} & \multicolumn{1}{l|}{11.45, 40.58} & \multicolumn{1}{l|}{82.66} & \multicolumn{1}{l|}{11.89, 97.54} & \multicolumn{1}{l|}{82.03} & \multicolumn{1}{l|}{11.87, 40.10} & \multicolumn{1}{l|}{78.76} & 18.09, 97.99 \\ 
\multicolumn{1}{|l|}{bge\_large\_pretrained\_finetuned}    & \multicolumn{1}{l|}{81.67} & \multicolumn{1}{l|}{10.34, 45.27} & \multicolumn{1}{l|}{80.48} & \multicolumn{1}{l|}{10.78, 97.26} & \multicolumn{1}{l|}{81.44} & \multicolumn{1}{l|}{73.0, 88.0}   & \multicolumn{1}{l|}{77.46} & 15.54, 98.35 \\ \hline
\multicolumn{1}{|l|}{llm\_embedder}                        & \multicolumn{1}{l|}{70.06} & \multicolumn{1}{l|}{10.83, 14.54} & \multicolumn{1}{l|}{68.26} & \multicolumn{1}{l|}{11.59, 96.83} & \multicolumn{1}{l|}{69.66} & \multicolumn{1}{l|}{11.59. 13.93} & \multicolumn{1}{l|}{68.58} & 20.5, 96.71  \\ 
\multicolumn{1}{|l|}{llm\_embedder\_pretrained}            & \multicolumn{1}{l|}{57.12} & \multicolumn{1}{l|}{5.42, 15.4}   & \multicolumn{1}{l|}{53.09} & \multicolumn{1}{l|}{5.94, 95.77}  & \multicolumn{1}{l|}{56.56} & \multicolumn{1}{l|}{47.0, 65.52}  & \multicolumn{1}{l|}{56.55} & 11.31, 95.77 \\ 
\multicolumn{1}{|l|}{llm\_embedder\_finetuned}             & \multicolumn{1}{l|}{81.58} & \multicolumn{1}{l|}{13.94, 22.1}  & \multicolumn{1}{l|}{82.28} & \multicolumn{1}{l|}{14.66, 97.34} & \multicolumn{1}{l|}{81.52} & \multicolumn{1}{l|}{14.63, 19.88} & \multicolumn{1}{l|}{79.14} & 20.73, 97.78 \\ 
\multicolumn{1}{|l|}{llm\_embedder\_pretrained\_finetuned} & \multicolumn{1}{l|}{80.37} & \multicolumn{1}{l|}{10.74, 25.01} & \multicolumn{1}{l|}{81.2}  & \multicolumn{1}{l|}{11.25, 97.32} & \multicolumn{1}{l|}{80.79} & \multicolumn{1}{l|}{11.23,23.22}  & \multicolumn{1}{l|}{79.44} & 15.82, 98.11 \\ \hline
\end{tabular}
}
\caption{Accuracy, $I_A$ and $I_B$ for  embeddings under different transformations.}
\label{tab:transformations}
\end{table*}

Correlation of $I_A$ and $I_B$ with accuracies across base, fine-tuned models with and without post-processing using transformations described in Section \ref{subsec:isotropyScores} is presented in Table \ref{tab:corr}. We see that, accuracy and both the isotropy scores are not correlated across datasets. Contrary to the conflicting claims in \cite{jung2023isotropic} and \cite{rudman2023stable}, our experiments establish that accuracy and isotropy scores are not correlated.

Combining these observations, we conclude that fine tuning improves the isotropy but isotropy cannot be attributed to retrieval accuracies. Our studies indicate that this may be the right resolution between the contradictions among studies by \cite{jung2023isotropic} and \cite{rudman2023stable} which we have discussed in Section \ref{sec:intro}.

\section{Recommendations and Conclusions}\label{sec:conclusions}

\subsection{Recommendations}\label{subsec:recc}

In this work, we have done a series of experiments to establish the impact of domain adaptation for embedding models. Based on this, we provide a set of recommendations to a researcher/practitioner on best using our findings. {\color{black} We provide anonymized code\footnote{\url{https://anonymous.4open.science/r/embedingStudy-E3B5/}} to perform the suggested steps, except domain adaptation, below}

\begin{itemize}
    \item Use a bootstrapped approach for obtaining accuracies as this will give not only point accuracies but also 95\% confidence intervals. 
    \item If possible, use domain adaptation - preferably pretraining followed by fine-tuning (PT-FT).
    \item Identify thresholds for the similarity scores - this will lead to bootstrapped accuracy which is statistically same as the full dataset bootstrapped accuracy, while suppressing less relevant documents  to end-users / downstream tasks.
    \item We propose two new metrics COE and ROE. The observed correlations, across 3 datasets, of the COE with accuracy and the ROE with thresholds indicate that they are reliable measures for the generalisation of performance on unseen data of that domain.
    \item Our results establish the lack of correlation of accuracies to isotropy scores. We thus suggest that computing isotropy scores to interpret retrieval accuracies is unlikely to be beneficial.
\end{itemize}

\subsection{Conclusions and Future Work} \label{sec:conclusion}
We have reported mean bootstrapped retrieval accuracies along with confidence intervals for various SOTA embedding models with and without domain-adaptation. We observe that fine-tuning (with or without pre-training) improves both mean and CI of retrieval accuracies. However, pre-training followed by fine-tuning improves CI further. 
We proposed a bootstrapped approach for choosing thresholds and observe that we can significantly reduce the number of retrieved sentences without any statistical deviation in retrieval performance. Our proposed cumulative distribution metrics, COE and ROE, to measure overlap between distributions of cosine similarities show strong correlations with retrieval performance and similarity thresholds respectively. We measure isotropy of embeddings using two independent SOTA isotropy metrics. We perform extensive evaluations on embeddings with and without isotropic transformations. We conclude that isotropy can be considered to be neither necessary nor sufficient from a retrieval accuracy perspective. Our study establishes systematic methods of analysing embeddings in specialised domains. The current work considers QA task only. Future work may involve other tasks like summarization, or multi-modal settings. 

\bibliographystyle{IEEEtran}
\bibliography{citation}

\end{document}